\documentclass[runningheads,a4paper]{llncs}

\usepackage{amssymb}
\usepackage{graphicx}
\usepackage{times}
\usepackage{helvet}
\usepackage{courier}
\usepackage{amssymb}
\usepackage{amsmath}
\usepackage{multirow}
\usepackage{amsopn}
\usepackage{algorithm}
\usepackage{algpseudocode}
\usepackage{epsfig}
\usepackage{graphicx,subfigure}
\usepackage{array}
\usepackage{xcolor}
\usepackage{url}
\usepackage{setspace}
\usepackage{url}
\usepackage{epstopdf}
\usepackage{marvosym}
\newcommand{\keywords}[1]{\par\addvspace\baselineskip
\noindent\keywordname\enspace\ignorespaces#1}

\begin{document}

\mainmatter  

\title{Learning beyond Predefined Label Space via Bayesian Nonparametric Topic Modelling}
\titlerunning{ }

\author{Changying Du$^{1,}$$^{2,}$$^{3}$\Letter         \and
        Fuzhen Zhuang$^1$        \and
        Jia He$^{1}$        \and
        Qing He$^1$              \and
        Guoping Long$^2$
}

\authorrunning{ }

\institute{   $^1$Key Lab of Intelligent Information Processing of Chinese Academy of Sciences (CAS),
              Institute of Computing Technology, CAS, Beijing 100190, China\\
              $^2$Laboratory of Parallel Software and Computational Science,
              Institute of Software, Chinese Academy of Sciences, Beijing 100190, China\\
              $^3$Huawei Noah's Ark Lab, Beijing 100085, China \\
              Email: duchangying@huawei.com
              }

%
%

\maketitle

\begin{abstract}
In real world machine learning applications, testing data may contain some meaningful new categories that have not been seen in labeled training data. To simultaneously recognize new data categories and assign most appropriate category labels to the data actually from known categories, existing models assume the number of unknown new categories is pre-specified, though it is difficult to determine in advance. In this paper, we propose a Bayesian nonparametric topic model to automatically infer this number, based on the hierarchical Dirichlet process and the notion of latent Dirichlet allocation. Exact inference in our model is intractable, so we provide an efficient collapsed Gibbs sampling algorithm for approximate posterior inference. Extensive experiments on various text data sets show that: (a) compared with parametric approaches that use pre-specified true number of new categories, the proposed nonparametric approach can yield comparable performance; and (b) when the exact number of new categories is unavailable, i.e. the parametric approaches only have a rough idea about the new categories, our approach has evident performance advantages.
\end{abstract}

\section{Introduction}

Human exploration of the world is never-ending, and we never know there still exist how many unknown things beyond our scope. For real-world machine learning applications, we often can only collect limited training instances before we do prediction on a large amount of unlabeled testing instances. Given the temporal and spatial constrictions at the beginning, it is likely that unlabeled new instances observed after a long time involve some meaningful new categories of objects, e.g., the news classification problem studied in \cite{zhang2011serendipitous,zhuang2010d,li2007learning}, and the bacterial detecting problem in \cite{akova2010machine,dundar2012bayesian}.

Basically, traditional classification models are unable to recognize new data categories, while clustering models cannot make full use of the supervised information from known categories.
An ideal model should simultaneously recognize the new data categories and assign most appropriate category labels to the data actually from known categories, since these two processes can benefit from each other.
Existing models for such a learning scenario typically assume the number of unknown new categories is pre-specified. In \cite{zhuang2010d}, Zhuang et al. proposed a double-latent-layered Latent Dirichlet Allocation (DLDA) model, which can utilize supervised information from known categories in a generative manner. While classifying test data into categories acquired from the training data, their model can simultaneously group the remaining data into some pre-specified number of new clusters. In \cite{zhang2011serendipitous}, the so-called Serendipitous Learning (SL) model established a maximum margin learning framework that combines the classification model built upon known classes with the parametric clustering model on unknown classes. Though these methods are effective when the true number of unknown new categories is available, their performances can be significantly degraded by a vague or wrong specification of the unknown category information.

Given that the accessibility assumption of the true number of unknown categories often is impractical, in this paper, we propose a Bayesian nonparametric topic model based on the hierarchical Dirichlet process \cite{teh2006hierarchical} and the notion of latent Dirichlet allocation \cite{blei2003latent}, for semi-supervised text modelling beyond the predefined label space. Unlike existing methods \cite{zhuang2010d,zhang2011serendipitous} which assume that the number of unknown new categories in test data is known, our model can automatically infer this number via nonparametric Bayesian inference while classifying the data from known categories into their most appropriate categories. Exact inference in our model is intractable, so we provide an efficient collapsed Gibbs sampling algorithm for approximate posterior inference.
Extensive experiments on various text data sets show that: (a) compared with parametric approaches that use pre-specified true number of new categories, the proposed nonparametric approach can yield comparable performance; and (b) when the exact number of new categories is unavailable, i.e. the parametric approaches only have a rough idea about the new categories, our approach has evident performance advantages.

In the following, we first review related works, and then present the generative process of our model and its approximate inference; experimental results are discussed in detail, before we conclude the paper and point out future work.

\section{Related Work}

A special case of the problem studied in this paper is the Positive and Unlabeled (PU) learning \cite{yu2003text,li2007learning,du2014analysis}, where the goal is to identify usually valuable positive instances from a huge collection of unlabeled ones. Our model generalizes PU learning in that, it not only identifies (multiple) known category of instances but also conducts nonparametric clustering for the remaining instances. It should be noted that the identification of known categories may benefit from a proper grouping of the unknown instance categories.

Assuming accessibility to both the seen and the unseen classes in the unlabeled data, the recently proposed Generalized Zero-Shot Learning (GZSL) \cite{NIPS2018_7471} is also related to our work. However, GZSL has to leverage semantic representations such as attributes or class prototypes to bridge seen and unseen classes, while our setting here is more challenging. Moreover, GZSL is not easy to infer the number of unseen classes underlying the data.

Another topic closely related to ours is semi-supervised clustering \cite{bilenko2004integrating}, which exploits available knowledge to help partition unlabeled data into groups. Generally, its knowledge is represented in the form of pairwise constraints \cite{bilenko2004integrating,kulis2009semi,rangapuram2012constrained}, i.e., cannot-link and must-link, which tends to be inefficient when the number of constraints is very large. Noting that our assumption is plenty of training instances are available from the known categories, these algorithms may suffer from efficiency problems. Moreover, violation of the constraints usually is allowed in these models, so it is not easy to map the resultant data clusters to the known classes. Instead of using constraints as supervision, we directly leverage label information in our model.

Under the nonparametric Bayesian framework, a semi-supervised determinantal clustering process was proposed in \cite{shah2013determinantal}. However, in each round of its sampling based inference procedure, its kernelized formulation leads to cubic computational complexity w.r.t. the number of instances to be clustered, which makes it infeasible for large data sets.

In nonparametric Bayesian statistics, the Dirichlet Process (DP) is a popular stochastic process that is widely used for adaptive modelling of the data \cite{teh2011dirichlet}. Intuitively, it is a distribution over distributions, i.e. each draw from a DP is itself a distribution.
Sethuraman \cite{sethuraman1991constructive} explicitly showed that distributions drawn from a DP are discrete with probability one, that is, the random distribution $G$ distributed according to a DP with concentration parameter $\gamma$ and base distribution $H$, can be written as
\begin{center}
$G =\sum_{i=1}^\infty\pi_i\delta_{\theta_i}$,\ \ \ \ $\pi_i=v_i\prod_{j=1}^{i-1}(1-v_j)$,
\end{center}
where $\theta_i\sim H$, $v_i\sim \text{Beta}(1,\gamma)$, and $\delta_\theta$ is an atom at $\theta$.
It is clear from this formulation that $G$ is discrete almost surely, that is, the support of $G$ consists of a countably infinite set of atoms, which are drawn independently from $H$.

Antoniak \cite{antoniak1974mixtures} first introduced the idea of using a DP as the prior for the mixing proportions of simple distributions, which is called the DP Mixture (DPM) model. Due to the fact that the distributions sampled from a DP are discrete almost surely, data generated from a DPM can be partitioned according to their distinct values of latent parameters $\theta_i$'s. Therefore, DPM is a flexible mixture model, in which the number of mixture components is random and grows as new data are observed.
Teh et al. \cite{teh2006hierarchical} proposed the Hierarchical DP (HDP), which is a nonparametric Bayesian approach to the modeling of grouped data, where each group is associated with a DPM model, and where we wish to link these mixture models.

\section{Learning beyond Predefined Labels via Generative Modelling}
\subsection{Problem specification}
Assume we have a labeled training data set $\mathcal{D}_l$ from the known categories $\mathcal{K}$,
and an unlabeled test data set $\mathcal{D}_u$ which includes instances from both the known categories $\mathcal{K}$ and some unknown new categories  $\mathcal{U}$.
The goal is to learn a function $f: \mathcal{D}_u \rightarrow \mathcal{K} \cup \mathcal{U}$ that maps any instance in $\mathcal{D}_u$ to its category label in $\mathcal{K} \cup \mathcal{U}$. Specifically, if an instance comes from the known categories $\mathcal{K}$, we aim to identify its true category label; meanwhile, we aim to group the instances not belonging to the known categories $\mathcal{K}$ into clusters $\mathcal{U}$.

\subsection{The Proposed Bayesian Nonparametric Topic Model}
For the problem specified above, an ideal model should simultaneously recognize the unknown new data categories and assign most appropriate category labels to the data actually from known categories, since these two processes can benefit from each other. However, it is usually difficult to determine the number of unknown categories in advance, which makes parametric approaches that assume this number is pre-specified impractical. To avoid performance degrading caused by a vague or wrong specification of the category information, in this paper, we propose a Bayesian nonparametric topic model, which can automatically infer the number of unknown new categories underlying test data $\mathcal{D}_u$ while classifying the data from known categories $\mathcal{K}$ into their most appropriate categories. Specifically, focusing on text data, we assume the following generative process for a document corpus:

\smallskip
\begin{enumerate}
\item Draw concentration parameters $\gamma \sim \Gamma(\gamma|a_\gamma,b_\gamma)$ and $\alpha \sim \Gamma(\alpha|a_\alpha,b_\alpha)$, where $a_\cdot$ and $b_\cdot$ are the shape and scale parameter of a Gamma distribution respectively;
\item Draw a discrete distribution $G_0 \sim \text{DP}(\gamma,H)$, where the base distribution $H$ is a $L$-dimensional Dirichlet with parameter $\zeta$, and $G_0$ has countable but infinite number of atoms;
\item Draw a discrete category distribution $G_d \sim \text{DP}(\alpha,G_0)$ for the $d$-th document;
\item Choose a document category $\varphi_{dn} \sim G_d$ for the $n$-th word in the $d$-th document\footnote{Note that, the support of the discrete distribution $G_d$ consists of atoms drawn from $G_0$, the atoms of which are eventually from $H$. Thus, $\varphi_{dn}$ is a vector rather than an index.};
\item Choose a word topic index $y_{dn} \sim \text{Categorical}(\varphi_{dn})$ for the $n$-th word in the $d$-th document;
\item Draw word topics $\phi_l \sim \text{Dir}(\beta),\ l=1,...,L$ from a $P$-dimensional Dirichlet prior with parameter $\beta$, where $L$ is the number of topics and $P$ is the vocabulary size;
\item Choose a word $w_{dn} \sim \text{Categorical}(\phi_{y_{dn}})$.
\end{enumerate}

\smallskip
As described above, this generative model integrates the hierarchical Dirichlet process (HDP) \cite{teh2006hierarchical} with the notion of Latent Dirichlet Allocation (LDA) \cite{blei2003latent}. However, the difference from standard LDA is that, here the distribution over word topics is conditioned on document categories rather than documents.
Placing a DP prior on the document category distribution $G_d$, a document is allowed to involve an infinite number of categories. Meanwhile, assuming multiple $G_d$'s have the same discrete base distribution $G_0$ (which also has a DP prior), multiple documents not only can have their distinct categories but also have the chance to share some common ones. The actual number of categories used to model a corpus is determined by nonparametric Bayesian posterior inference. Note that, if the category label of a document is known, we can fix the corresponding category of all words in this document to the category determined by the label during posterior inference. In this way, the supervision from known categories can be injected. For any document without known label,
we can infer the most appropriate category for each of its words, and assign this document to the category that generates most of its words.

\begin{figure}[!ht]
\begin{center}
\centerline{\includegraphics[width=0.6\columnwidth, height=3.3cm]{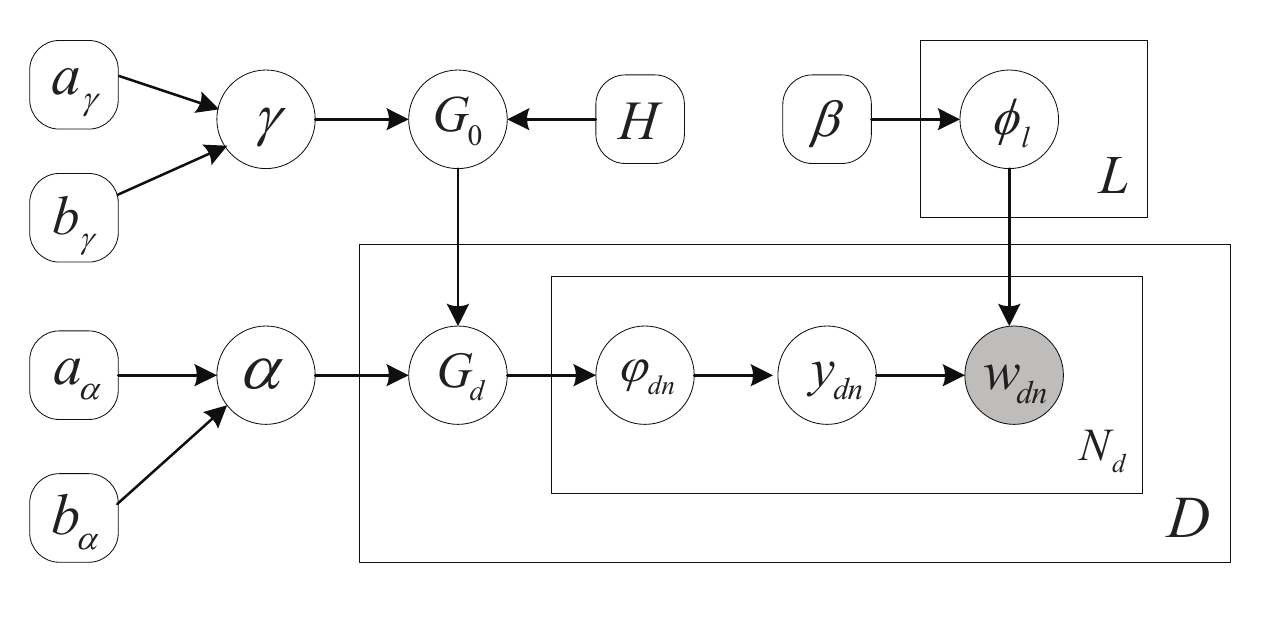}}
\caption{Graphical representation of the proposed LBPL-NTM model.
}
\label{HDP-SDC-model}
\end{center}
\end{figure}

Since the proposed model for Learning Beyond Predefined Labels (LBPL) is based on Nonparametric Topic Modelling (NTM), it will be denoted by LBPL-NTM in the sequel. The probabilistic generative process of LBPL-NTM is illustrated as a graphical model in Figure \ref{HDP-SDC-model}.

Note that, LBPL-NTM is conceptually different from the infinite extension of LDA presented in \cite{teh2006hierarchical}, which learns topics in a purely unsupervised manner and cannot make use of the labeled information. From pure modeling perspective, our model introduces an additional topic index layer ($y_{dn}$) along with $L$ hidden topics to infinite LDA. What's worth mentioning is that, it is not a trivial thing to extend the single-layered infinite LDA to a new two-layered model. With the introduced topic index layer and the hidden topics serving as low level topic modeling module, we can interpret $G_d$ as the distribution over categories (rather than over topics as in infinite LDA) for each document, and then inject labeled information through $\phi_{dn}$ and infer the number of unknown categories (rather than topics as in infinite LDA) automatically from the data.

Besides, LBPL-NTM also differs from supervised topic models \cite{mcauliffe2008supervised,zhu2012medlda,zhu2014gibbs} basically, which train discriminative classification models in the semantic space with pre-specified category labels and cannot identify new categories underlying the test data.

The labeled LDA model proposed in \cite{ramage2009labeled}, adopted a similar word-label correspondence idea by defining a one-to-one correspondence between LDA's latent topics and labels. However, it was designed to solve the multi-label problem in social bookmarking rather than discover new data categories underlying unlabeled data, thus is different from our model as well.

The double-latent-layered LDA (DLDA) \cite{zhuang2010d} is a more closely related work to ours, where the authors conditioned the distribution over word topics on the document categories as in our model. By utilizing supervised information from known categories in a generative manner, their parametric model can classify unlabeled data into categories acquired from the labeled data, while grouping data into some pre-specified number of new clusters simultaneously. Though DLDA is effective when the true number of new categories is available, its performance can be significantly degraded by a wrong specification of this number. Our key difference with theirs is that our nonparametric model can naturally deal with the scenario where the number of new categories underlying test data is not clear, via allowing an infinite number of categories to model the corpus.

For the model inference of LBPL-NTM, we need to compute the posterior distribution of hidden variables given the data and model hyper-parameters:
\begin{displaymath}
\begin{aligned}
p(\alpha,\gamma,\phi_l,\varphi_d,\mathbf{Y}_d|a_\alpha,b_\alpha,a_\gamma,b_\gamma,\beta,H,\mathbf{W}_d) \quad \quad \quad \quad \quad \\
=\frac{p(\alpha,\gamma,\phi_l,\varphi_d,\mathbf{Y}_d,\mathbf{W}_d|a_\alpha,b_\alpha,a_\gamma,b_\gamma,\beta,H)}{p(\mathbf{W}_d|a_\alpha,b_\alpha,a_\gamma,b_\gamma,\beta,H)}.
\end{aligned}
\end{displaymath}
However, the marginal probability in the denominator is intractable to compute. A popular way to conduct approximate posterior inference is the Markov Chain Monte Carlo (MCMC) method \cite{neal2000markov}. In the following, we will appeal to the Chinese restaurant franchise representation \cite{teh2006hierarchical} of HDP for approximate posterior sampling. Note that, the high-dimensional latent topics $\phi_l$'s and the latent category variables $\varphi_{dn}$'s are integrated out to attain efficient collapsed sampling.

\subsection{Inference by Collapsed Gibbs Sampling}

First we give a brief description of the Chinese restaurant franchise representation of HDP. In the Chinese restaurant franchise, the metaphor of the Chinese restaurant process is extended to allow multiple restaurants which share a set of dishes. A customer entering some restaurant sits at one of the occupied tables with a certain probability, and sits at a new table with the remaining probability. If the customer sits at an occupied table, he eats the dish that has already been ordered. If he sits at a new table, he needs to pick the dish for the table. The dish is picked according to its popularity among the whole franchise, while a new dish can also be tried.

To employ this representation of HDP for posterior sampling, we introduce necessary index variables.
Recall that $\varphi_{dn}$'s are random variables with distribution $G_d$.
Let $\theta_1,\cdots,\theta_K$ denote $K$ i.i.d. random variables (dishes) distributed according to $H$, and, for each $d$, let $\psi_{d1},\cdots,\psi_{dT_d}$ denote $T_d$ i.i.d. variables (tables) distributed according to $G_0$.
Then each $\varphi_{dn}$ is associated with one $\psi_{dt}$, while each $\psi_{dt}$ is associated with one $\theta_k$. Let $t_{dn}$ be the
index of the $\psi_{dt}$ associated with $\varphi_{dn}$, and let $k_{dt}$ be the index of $\theta_k$ associated with $\psi_{dt}$. Let $s_{dt}$
be the number of $\varphi_{dn}$'s associated with $\psi_{dt}$, $m_{dk}$ is the number of $\psi_{dt}$'s associated with $\theta_k$, and $m_k = \sum_{d}m_{dk}$ as the number of $\psi_{dt}$'s associated with $\theta_k$ over all $d$.

For each $d$, by integrating out $G_d$ and $G_0$, we have the following conditional distributions:
\begin{equation}
\renewcommand{\arraystretch}{1.5}
\begin{array}{rcl}
\varphi_{dn}|\varphi_{d1},\cdots,\varphi_{dn-1},\alpha,G_0\thicksim  \sum\limits_{t=1}^{T_d}\frac{s_{dt}}{n-1+\alpha}\delta_{\psi_{dt}}+\frac{\alpha}{n-1+\alpha}G_0,
\end{array}
\end{equation}
\begin{equation}
\renewcommand{\arraystretch}{1.5}
\begin{array}{rcl}
\psi_{dt}|\psi_{11},\psi_{12},\cdots,\psi_{21},\cdots,\psi_{dt-1},\gamma,H\thicksim \sum\limits_{k=1}^{K}\frac{m_{k}}{\sum_{k}m_k+\gamma}\delta_{\theta_{k}}+\frac{\gamma}{\sum_{k}m_k+\gamma}H.
\end{array}
\end{equation}

\smallskip
Note that, $t_{dn}$'s and $k_{dt}$'s inherit the exchangeability properties of $\varphi_{dn}$'s and $\psi_{dt}$'s, so the conditional distributions in (1) and (2) can be easily adapted to be expressed in terms of $t_{dn}$ and $k_{dt}$. In the following, we will alternately execute four steps: first sample $t_{dn}$ conditioned on all other variables, then sample $k_{dt}$ for each table of data, thirdly sample $y_{dn}$ for each word, and finally sample hyper-parameters $\gamma$ and $\alpha$. Note that, if the category label of a document is known, we fix the category index \textbf{$k$} of all words in this document to the label during the sampling process.

\vskip 0.05in
\textbf{Sampling $\mathbf{t}$.}$\ $ To compute the conditional distribution of $t_{dn}$ given the remaining variables,
we make use of exchangeability and treat $t_{dn}$ as the last variable being sampled in the last group. 
Using (1), the prior probability that $t_{dn}$ takes on a particular previously seen value $t$ is proportional
to $s^{-dn}_{dt}$ , whereas the probability that it takes on a new value (say $t^{new} = T_j +1$) is proportional
to $\alpha$. The likelihood of the data given $t_{dn}= t$ for some previously seen $t$ is simply $f(y_{dn}|\theta_{k_{dt}})$.
To determine the likelihood when $t_{dn}$ takes on value $t^{new}$, the simplest approach would be to generate
a sample for $k_{dt^{new}}$ from its conditional prior (2) \cite{neal2000markov}. If this value of $k_{dt^{new}}$ is itself a new value, say $k^{new} = K + 1$, we may generate a sample for $\theta_{k^{new}}$ as well.

Combining all this information, the conditional posterior distribution of $t_{dn}$ is then
\begin{equation} \label{crp2}
p(t_{dn}=t|\mathbf{t}^{-dn},\mathbf{k},\mathbf{Y},\Theta)\propto
\left\{
\begin{array}{@{}l@{\ \ \ }l}
    \alpha f(y_{dn}|\theta_{k_{dt}}), & t = t^{\text{new}},\\
    s_{dt}^{-dn}f(y_{dn}|\theta_{k_{dt}}), & t\  \text{appeared}.\\
\end{array}
\right .
\end{equation}

However, here we show that we don't need to store and update the $\theta$'s, i.e., we can get a collapsed sampler.
To compute the likelihood that $y_{dn}$ comes from the $k$-th class $\theta_k$, $1\leq k\leq K$, we can first compute the posterior distribution of $\theta_{k}$ given $\mathbf{Y}_{(k)}^{-dn}$ (elements assigned to class $k$ in $\mathbf{Y}^{-dn}$), then integrate over this posterior. Specifically, by conjugacy the posterior of $\theta_{k}$ is also Dirichlet distributed, whose parameter is updated from the prior base distribution $H$ according to $\mathbf{Y}_{(k)}^{-dn}$. If we assume $O_{kl}$ is the number of elements in $\mathbf{Y}_{(k)}^{-dn}$ that equal to $l,\ 1\leq l\leq L$, then
\begin{displaymath} \label{crp2}
\begin{aligned}
\theta_k|H,\mathbf{t}^{-dn},\mathbf{k}^{-dt_{dn}},\mathbf{Y}^{-dn}\sim \text{Dir}(\zeta+O_{k\cdot}),
\end{aligned}
\end{displaymath}
\vskip 0.05in
\noindent where $\zeta$ and $O_{k\cdot}$ both are $L$ dimensional vectors. Integrate over this posterior we can get the likelihood for $y_{dn}$,
\begin{equation} \label{crp2}
\begin{aligned}
f(y_{dn}|\theta_k:1\leq k\leq K)=\int \theta_{y_{dn}} \cdot \text{Dir}(\theta;\zeta+O_{k\cdot}) d\theta
=\frac{\zeta_{y_{dn}}+O_{ky_{dn}}}{\sum_l(\zeta_l+O_{kl})}.
\end{aligned}
\end{equation}

To compute the likelihood that $y_{dn}$ comes from a new $k=(K+1)$-th class $\theta_{K+1}$, we can directly integrate over the prior $H$:
\begin{equation} \label{crp2}
\begin{aligned}
f(y_{dn}|\theta_k:k=K+1)=\int \theta_{y_{dn}} \cdot \text{Dir}(\theta;\zeta) d\theta=\frac{\zeta_{y_{dn}}}{\sum_l\zeta_l}.
\end{aligned}
\end{equation}

\textbf{Sampling $\mathbf{k}$.}$\ $ Sampling the variables $k_{dt}$ is similar to sampling $t_{dn}$. Since changing $k_{dt}$ actually changes the component membership of all data items in table $t$, the likelihood of setting $k_{dt} = k$ is given by $\prod_{n:t_{dn}=t} f(y_{dn}|\theta_k)$, so that the conditional probability of $k_{dt}$ is
\begin{equation} \label{crp2}
\begin{aligned}
p(k_{dt} =k|\mathbf{t},\mathbf{k}^{-dt},\mathbf{Y},\Theta)\propto
\left\{
  \begin{array}{@{}l@{\ \ \ }l}
    \gamma \prod_{n:t_{dn}=t} f(y_{dn}|\theta_k), & k = k^{\text{new}},\\
    m_{k}^{-dt}\prod_{n:t_{dn}=t} f(y_{dn}|\theta_k), & k\  \text{appeared},\\
  \end{array}
\right .
\end{aligned}
\end{equation}
\noindent where $f(y_{dn}|\theta_k)$ can be computed same as above.

\vskip 0.15in
\textbf{Sampling $\mathbf{Y}$.}$\ \ $
Conditioned on $\mathbf{t}$, $\mathbf{k}$ and $\mathbf{Y}^{-dn}$, the prior of $y_{dn}=l,\ 1\leq l\leq L$ is:
\begin{eqnarray}
\nonumber p(y_{dn}=l|\mathbf{t},\mathbf{k}, \mathbf{Y}^{-dn})=\int \theta_{l} \cdot \text{Dir}(\theta;\zeta+O_{k_{dt_{dn}}\cdot}) d\theta  =\frac{\zeta_{l}+O_{k_{dt_{dn}}l}}{\sum_l(\zeta_l+O_{k_{dt_{dn}}l})}.
\end{eqnarray}

Assume $\mathbf{W}_{(l)}^{-dn}$ denotes the elements in $\mathbf{W}^{-dn}$ that are generated from topic $l$, and $O_{lw}$ is the number of elements in $\mathbf{W}_{(l)}^{-dn}$ that equal to $w,\ 1\leq w\leq P, \ 1\leq l\leq L$, then
\begin{displaymath} \label{crp2}
\begin{aligned}
\phi_l|\beta,\mathbf{Y}^{-dn},\mathbf{W}_{(l)}^{-dn}\sim \text{Dir}(\beta+O_{l\cdot}),
\end{aligned}
\end{displaymath}
\noindent where $\beta$ and $O_{l\cdot}$ both are $P$ dimensional vectors. Integrating over this posterior, we can get the likelihood that $w_{dn}$ is generated from topic $\phi_{l}$:
\begin{eqnarray}
\nonumber f(w_{dn}|\mathbf{t},\mathbf{k}, \mathbf{Y}^{-dn}, \mathbf{W}^{-dn})  =  \int \phi_{w_{dn}} \cdot \text{Dir}(\phi;\beta+O_{l\cdot}) d\phi =  \frac{\beta_{w_{dn}}+O_{lw_{dn}}}{\sum_w(\beta_w+O_{lw})}.
\end{eqnarray}

\vskip 0.05in

The conditional posterior probability of $y_{dn}=l,\ 1\leq l\leq L$ is proportional to the prior times the likelihood:
\begin{equation} \label{crp2}
\begin{aligned}
p(y_{dn}=l|\mathbf{t},\mathbf{k}, \mathbf{Y}^{-dn}, \mathbf{W})\propto
\frac{\zeta_{l}+O_{k_{dt_{dn}}l}}{\sum_l(\zeta_l+O_{k_{dt_{dn}}l})}\cdot \frac{\beta_{w_{dn}}+O_{lw_{dn}}}{\sum_w(\beta_w+O_{lw})}.
\end{aligned}
\end{equation}

\vskip 0.05in
\textbf{Sampling $\gamma$ and $\alpha$.}$\ $ In each iteration of our Gibbs sampling, we use the auxiliary variable method described in \cite{teh2006hierarchical} to sample $\gamma$ and $\alpha$.

\vskip 0.1in
We summarize the above approximate posterior sampling process in Algorithm 1. After this sampling process converges, we take a sample from the Markov chain and count the words assigned to each category $k=1,2,...$ for each document, and finally a document is assigned to the category that has generated most of its words.

\begin{algorithm} [ht] \small
\caption{Collapsed Gibbs Sampling for LBPL-NTM}
\vspace{0.03in}
\textbf{Input:} the words $\textbf{W}$, the number of topics $L$, parameter $\zeta$ of the base Dirichlet distribution $H$, the hyper-parameters $\beta$, $a_\gamma$, $b_\gamma$, $a_\alpha$, $b_\alpha$, and the maximal number of iterations $maxIter$.
\\
\textbf{Output:} $\textbf{t}$, $\textbf{k}$ and $\textbf{Y}$.
\begin{enumerate}
   \item Initialize the latent variables $\textbf{t}$, $\textbf{k}$, $\textbf{Y}$, $\gamma$ and $\alpha$;
   \item \textbf{for} $iter=1$ to $maxIter$ \textbf{do}
   \item   \ \quad Update $\textbf{t}$ according to (3), (4), and (5);
   \item   \ \quad Update $\textbf{k}$ according to (6), (4), and (5);
   \item   \ \quad Update $\textbf{Y}$ according to (7);
   \item   \ \quad Update $\gamma$ and $\alpha$ using the auxiliary variable method in \cite{teh2006hierarchical};
   \item \textbf{end for}
   \item Output $\textbf{t}$, $\textbf{k}$ and $\textbf{Y}$.
\end{enumerate}
\end{algorithm}

\subsection{Computational complexity}

In each round of our collapsed Gibbs sampling, the dominant computation is $O(|\textbf{W}_u|\cdot(|\bar{\mathbf{t}}|+|\mathbf{k}|) + |\textbf{W}_a|\cdot L)$, where $|\textbf{W}_u|$ is the total number of words in the unlabeled documents, $|\textbf{W}_a|$ is the total number of words in the entire corpus, $|\bar{\mathbf{t}}|$ is the average number of inferred word groups in each document, $|\mathbf{k}|$ is the inferred number of categories, and $L$ is the specified number of topics. Generally, $|\mathbf{k}|$ and $|\bar{\mathbf{t}}|$ are very small, and $L=128$ throughout the paper\footnote{For fair comparison with the DLDA model \cite{zhuang2010d}, the number of topics $L$ is fixed to the constant 128. We empirically find that $L$ has little performance influence (compared to the number of categories) on the learning problem studied here, as long as it is not too small or too large. This is probably due to the two-layered nature of our model.}, thus our model can be seen as scale linearly with the number of words in the corpus.
\vspace{0.03in}


\section{Experiments}
In this section, we evaluate the proposed LBPL-NTM model on various text corpora, including the benchmark 20 Newsgroups data set, the imbalanced TDT2 data set and the sparse ODP data set.

\subsection{Baselines and evaluation metrics}
We compare LBPL-NTM with the following algorithms:
\begin{itemize}
\item Serendipitous Learning (SL) \cite{zhang2011serendipitous}: a maximum margin learning framework that combines the classification model built upon known classes and the parametric clustering model on unknown classes;
\item DLDA \cite{zhuang2010d}: a double-latent-layered LDA model, which can utilize supervised information similar as LBPL-NTM when clustering data with pre-specified number of clusters;
\item Constrained 1-Spectral Clustering (COSC) \cite{rangapuram2012constrained}: a state-of-the-art graph-based constrained clustering algorithm, which can guarantee that all given constraints are fulfilled;
\item Semi-supervised K-means (SSKM) \cite{kulis2009semi}: clustering data with pairwise constraints in original space;
\item Unsupervised clustering package CLUTO\footnote{http: //glaros.dtc.umn.edu/gkhome/cluto/cluto/download};
\item Nonparametric Bayesian unsupervised clustering model Dirichlet Process Gaussian Mixture (DPGM).

\end{itemize}

\smallskip
Two popular clustering metrics are adopted to compare the clustering quality of these algorithms: normalized mutual information (NMI) \cite{manning2008introduction} and adjusted rand index (ARI) \cite{hubert1985comparing}.
NMI measures how closely the clustering algorithm could reconstruct the label distribution underlying the data. If $A$ and $B$ represent the cluster assignments and the ground truth class assignments of the data respectively, then NMI is defined as
\begin{displaymath}
NMI = 2\cdot I(A;B)/(H(A)+H(B)),
\end{displaymath}
where $I(A;B)=H(A)-H(A|B)$ is the mutual information between $A$ and $B$, $H(\cdot)$ is the Shannon entropy, and $H(A|B)$ is the conditional entropy of $A$ given $B$.

If $a$ denotes the number of pairs of data points that are in the same cluster in $A$ and in the same class in $B$, and $b$ denotes the number of pairs of points that are in different clusters in $A$ and in different classes in $B$, then the Rand Index (RI) is given by $RI = (a + b)/C_2^D$, where $C_2^D$ is the total number of possible pairs in the dataset. Since the expected RI value of two random assignments does not take a constant value, Hubert and Arabie \cite{hubert1985comparing} proposed to discount the expected RI of random assignments by defining the ARI as
\begin{displaymath}
ARI = (RI-Expected\_RI)/(\max(RI)-Expected\_RI).
\vspace{0.05in}
\end{displaymath}

As in \cite{zhuang2010d}, we also evaluate the classification accuracy on the data from the known classes with average $F1$ measure.
For each known class, the $F1$ score can be computed as follows,
\begin{displaymath}
F1_i = 2\cdot Precison_i \cdot Recall_i/(Precison_i + Recall_i), \ i=1,...,k,
\end{displaymath}
where $Precison_i$ and $Recall_i$ are the precision and recall on the $i$-th known class. Then, we use the average $F1$ score over these $k$ known classes as the final measure.

\subsection{Parameter settings}

In all our experiments, we set the parameters and hyper-parameters of LBPL-NTM as follows: $L=128$, $a_\gamma=1, b_\gamma=0.001$, $a_\alpha = 5$, $b_\alpha = 0.1$, $\zeta_l=1,\ l=1,...,L$, $\beta_w=0.01,\ w=1,...,P$. We run 3000 Gibbs sampling iterations to sample from the posteriors of LBPL-NTM and DLDA, and use the last sample for classification and clustering performance evaluation\footnote{Such a choice is consistent with the evaluation strategy in \cite{zhuang2010d}. Alternatively, we can also average the classification and clustering scores over multiple posterior samples.}.

The parameter settings of all compared algorithms follow the instructions in their original papers and are carefully tuned on our data sets. The similarity matrix for COSC is constructed using the cosine value of the angle between each pair of documents\footnote{COSC works not well with the $k$-NN similarity graph \cite{buhler2009spectral} on our data sets.}. For CLUTO, we use its \emph{direct} implementation for clustering with default parameter settings. PCA is used to reduce the original high dimensionality to 500 for SL and DPGM, due to efficiency problems. Without statement, all algorithms except for DPGM and LBPL-NTM, use the true number of data categories.

\subsection{Evaluation results}

\textbf{Benchmark data---20 Newsgroups:}
This data set is widely used in text categorization and clustering. It has approximately 20,000 newsgroup documents that are evenly partitioned into twenty different newsgroups. Since some of the newsgroups are very closely related, a part of these twenty newsgroups are further grouped into four top categories, e.g., the top category \emph{sci} contains four subcategories \emph{sci.crypt}, \emph{sci.electronics}, \emph{sci.med} and \emph{sci.space}. We only retain the terms that have document frequency (DF) above 15 and are not in the stop words list. As in Table II of \cite{zhuang2010d}, we consider two kinds of 4-way learning problems---the data for each \emph{difficult} problem consist of all 4 subcategories of a top category, and the data for each \emph{easy} problem consist of 4 subcategories from different top categories. Here these problems are denoted as E1-E4 and D1-D4 for short. For each problem, assume we have supervision from the subcategories in bold face in Table II of \cite{zhuang2010d}, from which 40\% instances are sampled as training data, and the rest 60\% and all instances from the subcategories without supervision are used as testing data. We independently repeat the experiments 10 times, and the averaged results over these trials are reported in Figure 2, from which we can see LBPL-NTM and DLDA can significantly outperform other competitors, while these two methods perform similarly. However, it should be noted that DLDA used the actual number of categories, while LBPL-NTM can automatically infer the most appropriate number from data owing to the merits of Bayesian nonparametrics. The posterior frequencies of the inferred numbers of categories by LBPL-NTM are shown in Figure 3, from which we can see higher frequencies around the true number 4.

One may naturally question the learning performance of DLDA when actual number of categories is not available. To this end, we further compare LBPL-NTM with DLDA, assuming that we only have a rough idea about the number of unknown categories underlying data. Under the same settings as above, Figure 4 gives the average results over 10 independent trials on 20 Newsgroups data set when the number of categories $K$ in DLDA is varied from $K=3$ to $K=7$ (the true number is 4). From these results we can observe that 1) the clustering performance (in terms of NMI and ARI) of DLDA is quite sensitive to the pre-specified $K$ while LBPL-NTM can circumvent this issue with nonparametric prior; 2) it seems that the classification performance (F1) of DLDA becomes better when the specified number of categories is larger, but as will be seen later this is not always true.

\smallskip
\textbf{Imbalanced data---TDT2:}
The NIST Topic Detection and Tracking (TDT2) corpus consists of data collected during the first half of 1998 and taken from 6 sources, including 2 news wires, 2 radio programs and 2 television programs. It consists of 11201 on-topic documents which are classified into 96 semantic categories. In the experiment, those documents appearing in two or more categories were removed, and only the largest 20 categories were kept. As above, we only retain the terms that have DF above 15 and are not in the stop words list. Here we assume supervision is available in the largest 10 categories, from which 40\% instances are sampled as training data, and the rest 60\% and all instances from the categories without supervision are used as testing data.
We independently repeat the experiments 10 times, and the averaged results over these trials are shown in Table 1. It seems that the parametric approach DLDA doesn't get its best performance when the true number of categories is pre-specified, which is probably due to the severe imbalance among different categories. Surprisingly, LBPL-NTM achieves the best results without any information of the total number of data categories. This may be due to its ability to dynamically adjust the number of data categories during its posterior sampling process. The posterior frequencies of the inferred numbers of categories by LBPL-NTM are shown in Figure \ref{NIPSresults11a}.

\begin{figure}[H]
\centering
\subfigure[NMI]{
    \includegraphics[height=1.5in,width=4.3in]{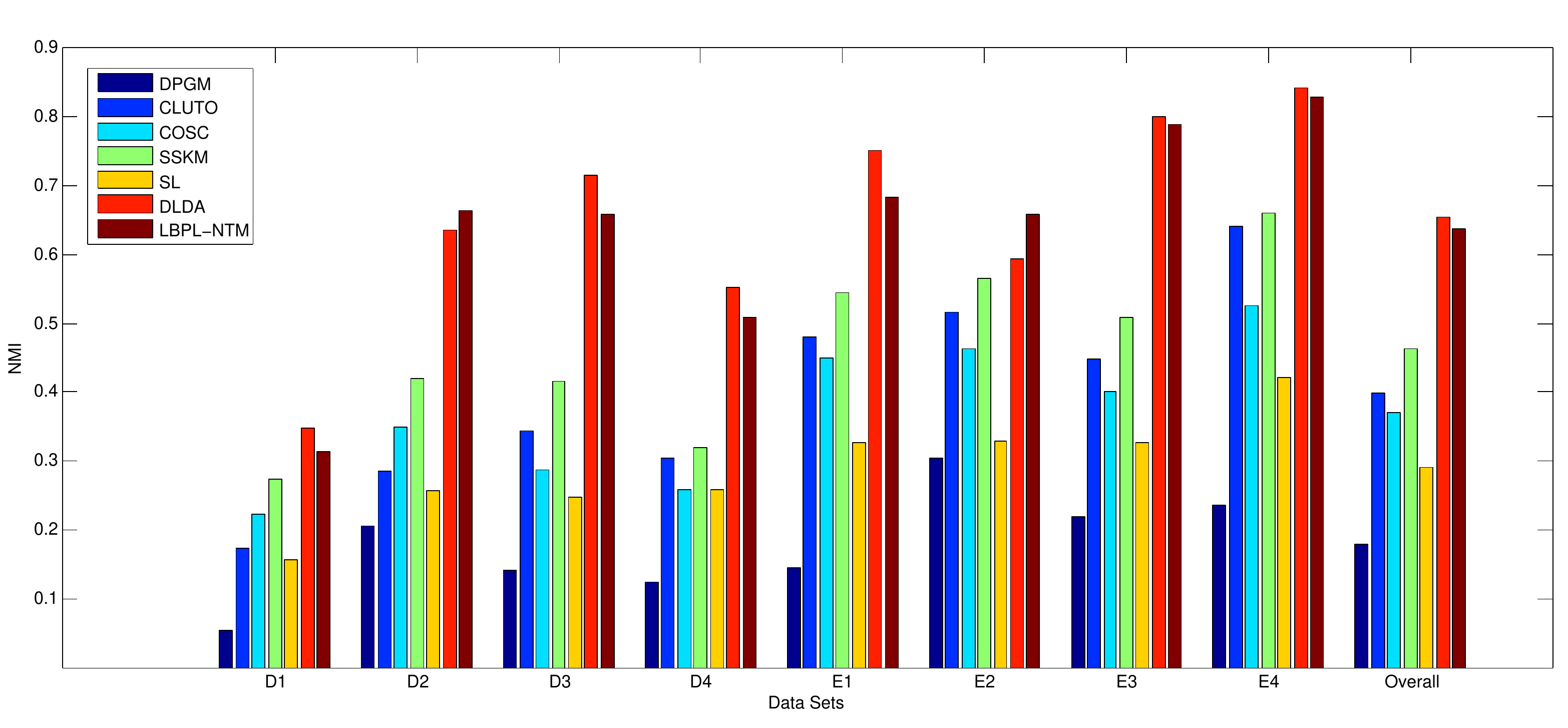}}
\subfigure[ARI]{
    \includegraphics[height=1.5in,width=4.3in]{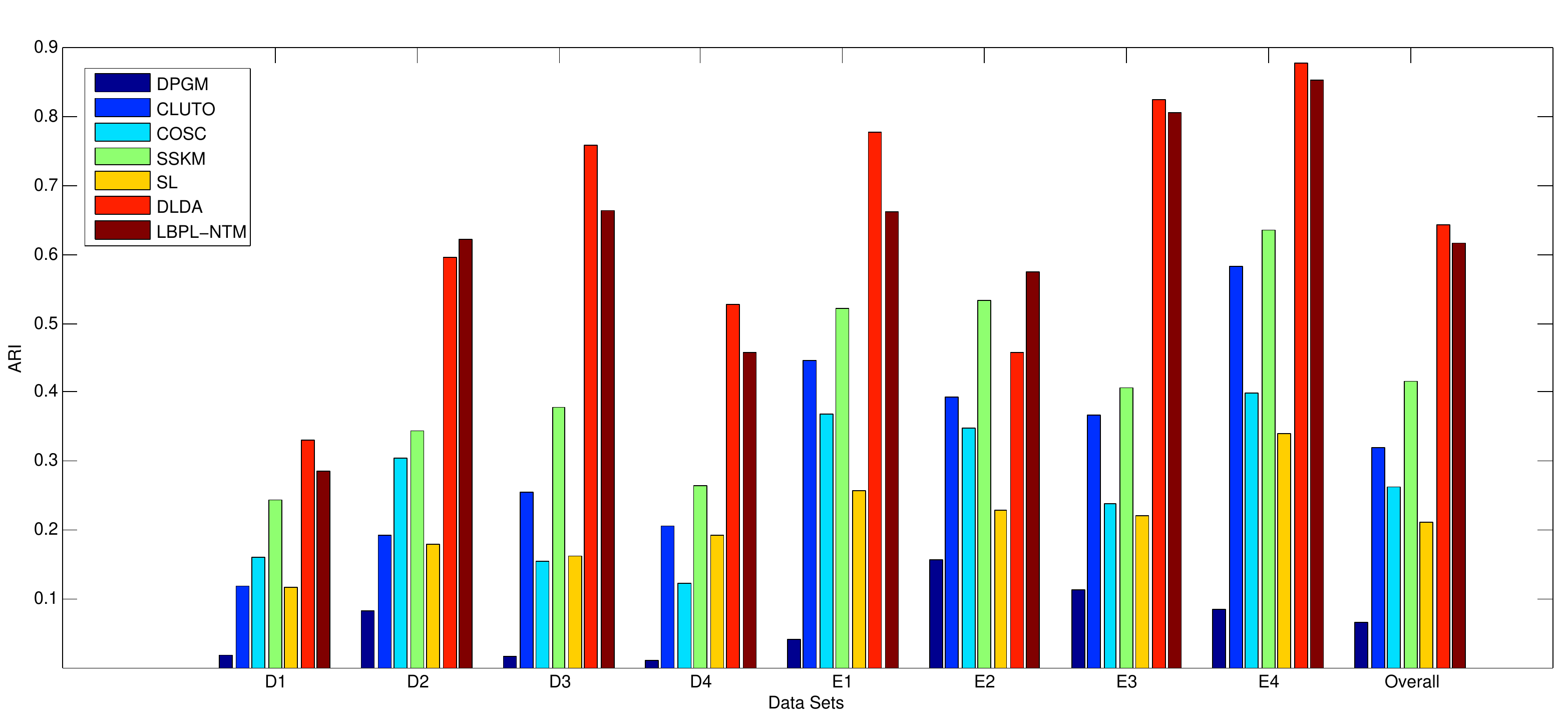}}
\subfigure[F1]{
    \includegraphics[height=1.5in,width=4.3in]{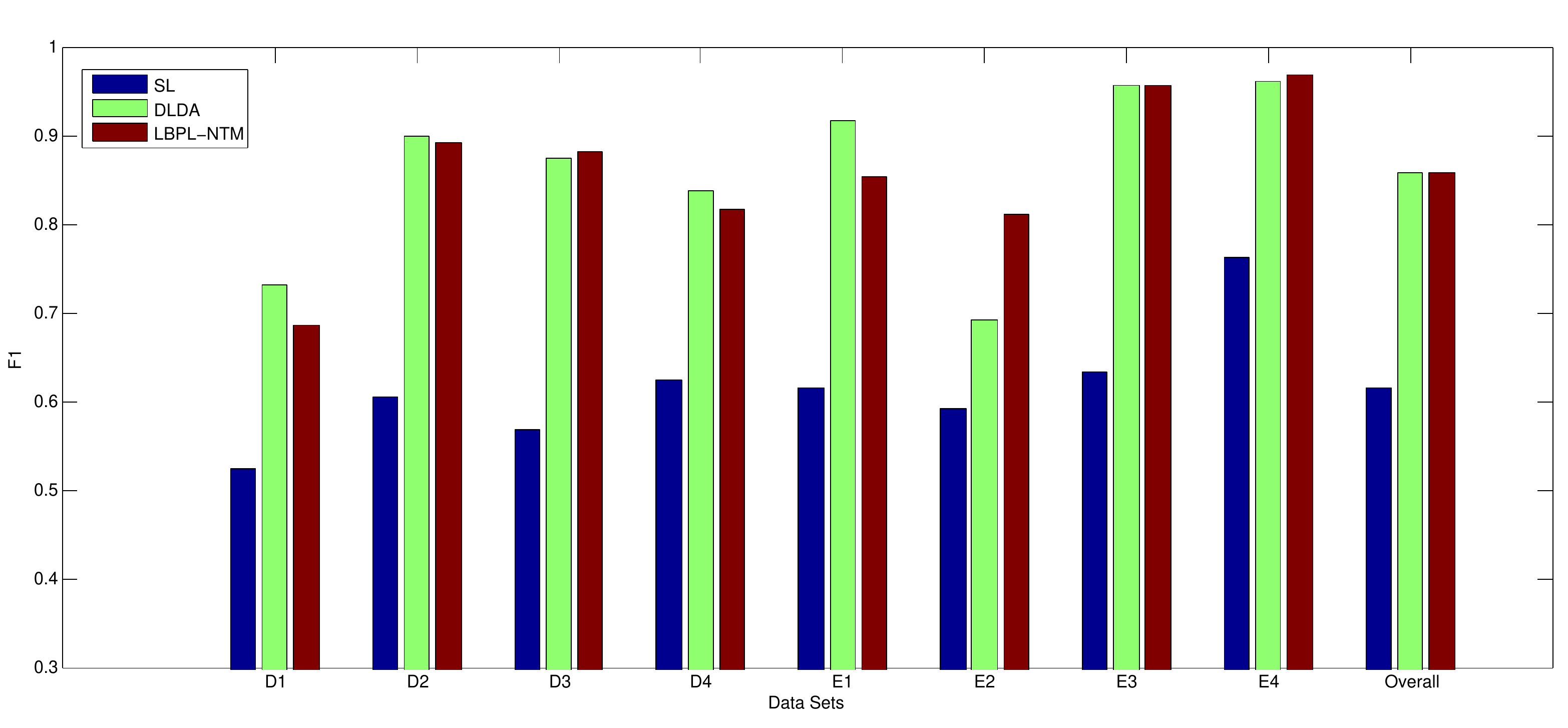}}
  \caption{Comparison on the 4-way learning problems (D1-D4, E1-E4) constructed from 20 Newsgroups data. All results are averaged over 10 independent trials in terms of NMI, ARI and F1.}
\vskip-0.1in
\end{figure}

\begin{figure}[H]
\centering
{\includegraphics[height=0.88in,width=2.1in]{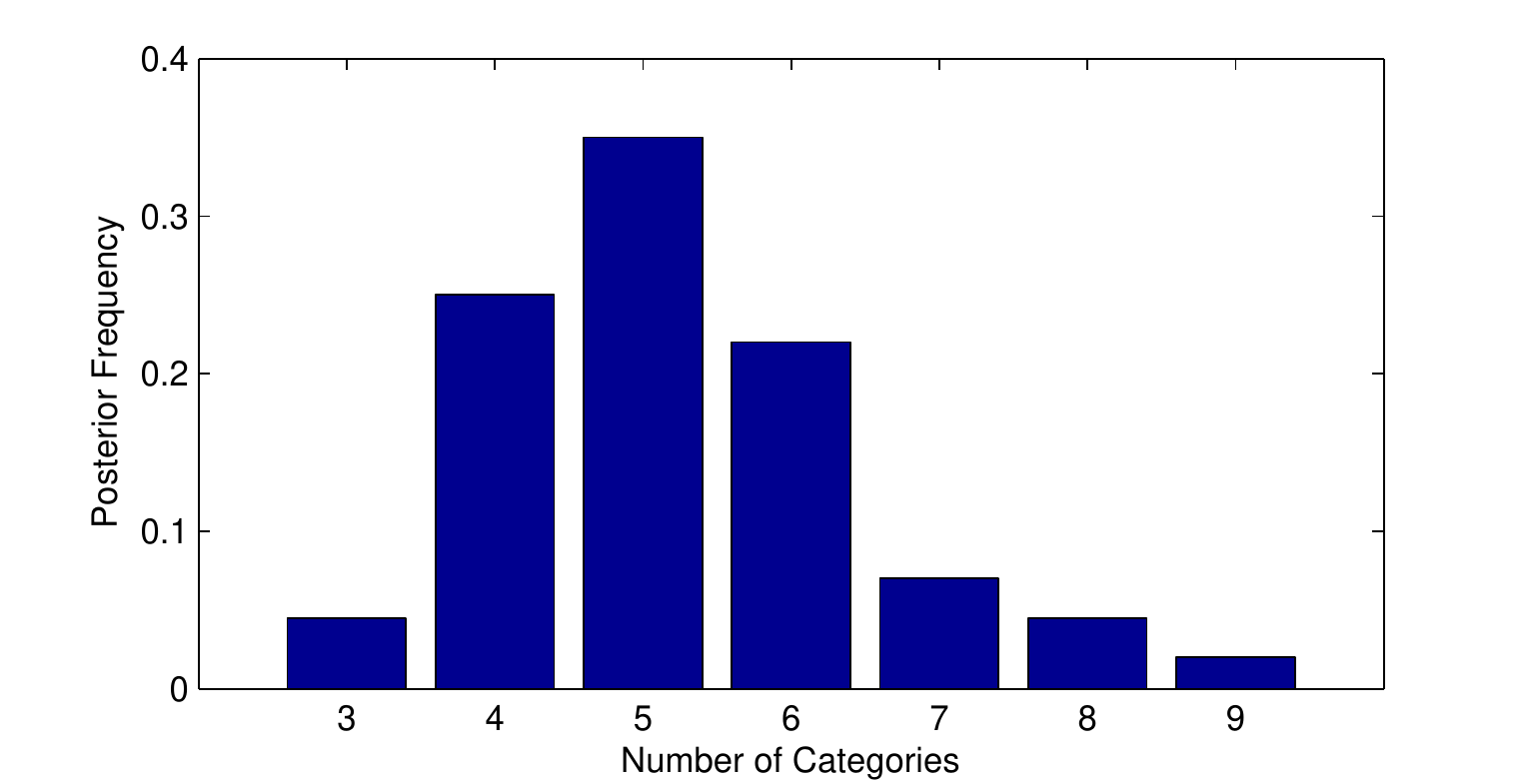}}
\caption{Posterior frequencies of the inferred numbers of categories on 20 Newsgroups.}
\end{figure}

\begin{table}[!htbp]
\scriptsize
\centering
\setlength\tabcolsep{6.5pt}
\renewcommand\arraystretch{1.1}
\caption{Averaged results over 10 independent trials on the TDT2 data.}
\label{NIPSresult8}
\begin{tabular}{c|c|c|c|c|c|c|c|c|c}
\hline
& \multirow{2}{*}{DPGM}   & \multirow{2}{*}{CLUTO}  &\multirow{2}{*}{COSC}  & \multirow{2}{*}{SSKM} & \multirow{2}{*}{SL}  & \multicolumn{3}{c|}{DLDA} & \multirow{2}{*}{LBPL-NTM}\\
  \cline{7-9}
&&&&&&$K$=15&$K$=20&$K$=25&\\
\hline
NMI  & 0.4878     & 0.8217  & 0.6042   & 0.8057 & 0.7743  & 0.8157     & 0.8173    & 0.8135     & \textbf{0.8358}  \\
ARI  & 0.2608     & 0.6591  & 0.4375   & 0.6665 & 0.7159  & 0.7804     & 0.7167    & 0.6788     & \textbf{0.7873}  \\
F1   & -          & -       & -        & -      & 0.8443  & 0.8473     & 0.8490    & 0.8068     & \textbf{0.9075}  \\
\hline
\end{tabular}
\vskip-0.1in
\end{table}

\begin{figure}[H]
\centering
\subfigure[NMI]{
      \includegraphics[height=1.6in,width=4.3in]{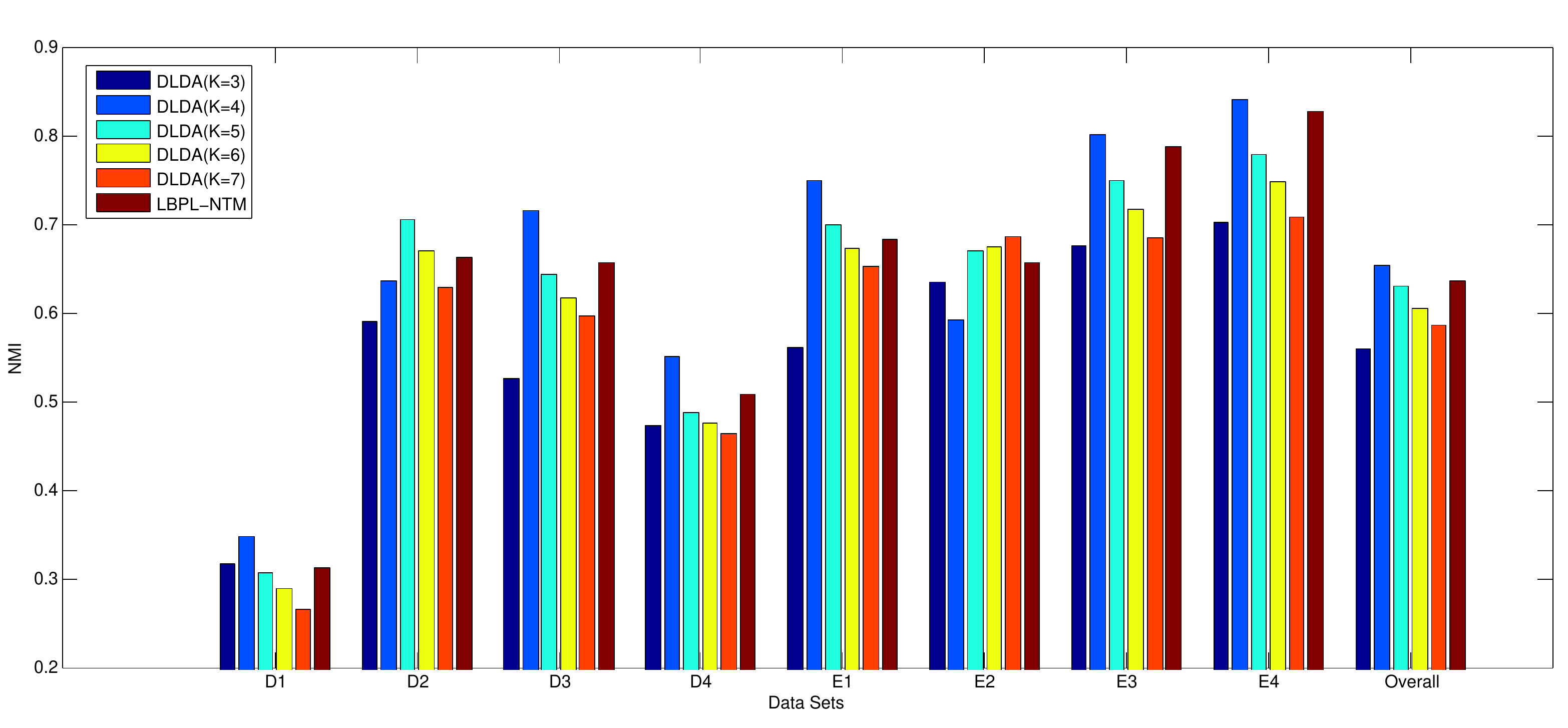}}
\subfigure[ARI]{
      \includegraphics[height=1.6in,width=4.3in]{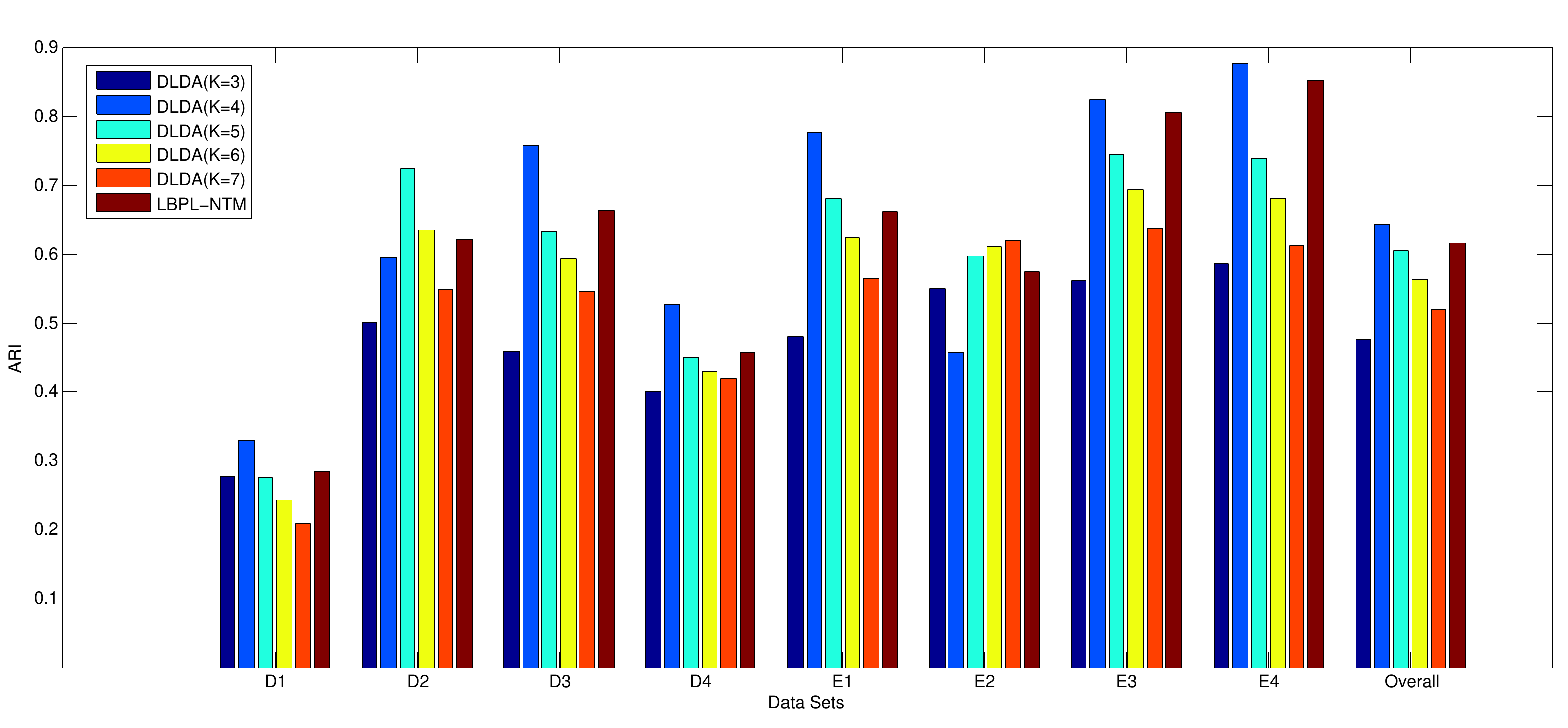}}
\subfigure[F1]{
      \includegraphics[height=1.6in,width=4.3in]{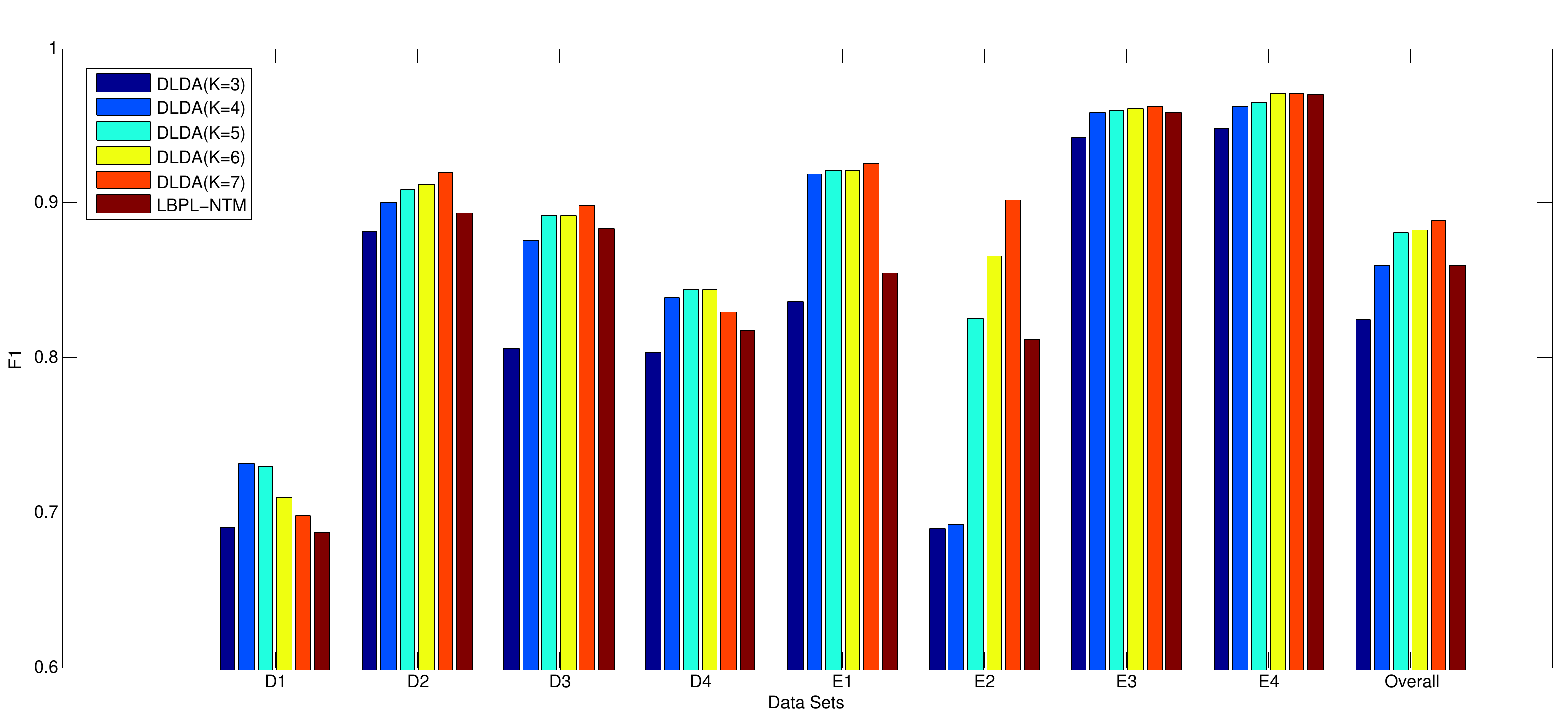}}
\caption{Further comparison with DLDA on the 4-way learning problems constructed from 20 Newsgroups. All results are averaged over 10 independent trials in terms of NMI, ARI and F1.}
\end{figure}

\smallskip
\textbf{Sparse data---ODP:}
This data set is collected by Yin et al. \cite{yin2009exploring}, originally for web object classification by exploiting social tags. It contains 5536 web pages from 8 categories, which are detailed in Table 1 in \cite{yin2009exploring}. Since the features on each web page are the social tags on it, these data are extremely sparse. Specifically, the average number of tag words on each web page is 25.76, which is much smaller than that (more than 160) of 20 Newsgroups. Assume that there is supervised information in the categories of Books, Electronic, Health and Garden. As above, we randomly sample 40\% instances as training data from these known categories, and the rest 60\% and all instances from the categories without supervision are used as testing data. We independently repeat the experiments 10 times, and report the averaged NMI, ARI and F1 values in Table \ref{NIPSresult10}, from which we can see LBPL-NTM also has competitive performance on sparse data. The posterior frequencies of the inferred numbers of categories are shown in Figure \ref{NIPSresults11b}. Note that there is a very small category---Office in ODP, and it is not easy to discover it due to data sparseness.

\begin{table}[!htbp]
\scriptsize
\centering
\setlength\tabcolsep{6.5pt}
\renewcommand\arraystretch{1.2}
\caption{Averaged results over 10 independent trials on the ODP data.}
\label{NIPSresult10}
\begin{tabular}{c|c|c|c|c|c|c|c|c|c}
\hline
& \multirow{2}{*}{DPGM}   & \multirow{2}{*}{CLUTO}  &\multirow{2}{*}{COSC}  & \multirow{2}{*}{SSKM} & \multirow{2}{*}{SL}  & \multicolumn{3}{c|}{DLDA} & \multirow{2}{*}{LBPL-NTM}\\
  \cline{7-9}
&&&&&&$K$=5&$K$=8&$K$=10&\\
\hline
NMI  & 0.2825     & 0.5302  & 0.3866   & 0.5155 & 0.4523  & 0.6039     & 0.6054    & \textbf{0.6084}     &0.5877  \\
ARI  & 0.0966     & 0.3983  & 0.2451   & 0.4145 & 0.3684  & \textbf{0.6034}     & 0.5480    & 0.5119     &0.5781  \\
F1   & -          & -       & -        & -      & 0.7045  & 0.7400     & \textbf{0.7868}    & 0.7461     &0.7715  \\
\hline
\end{tabular}
\end{table}

\begin{figure}[H]
\centering
\subfigure[TDT2]{
\label{NIPSresults11a}
 {\includegraphics[height=1in,width=2.18in]{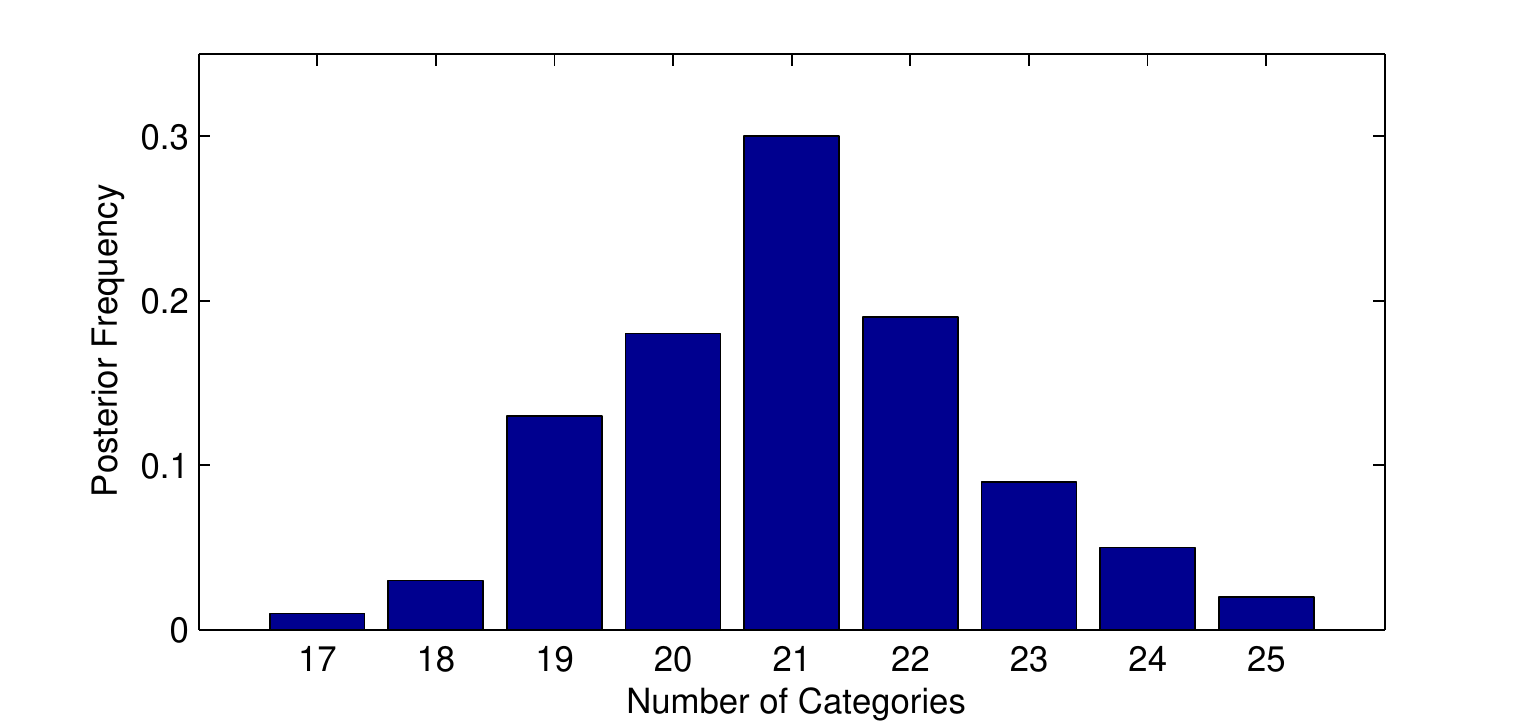}}}
\hskip 0.1in
\subfigure[ODP] {
\label{NIPSresults11b}
{\includegraphics[height=1in,width=1.98in]{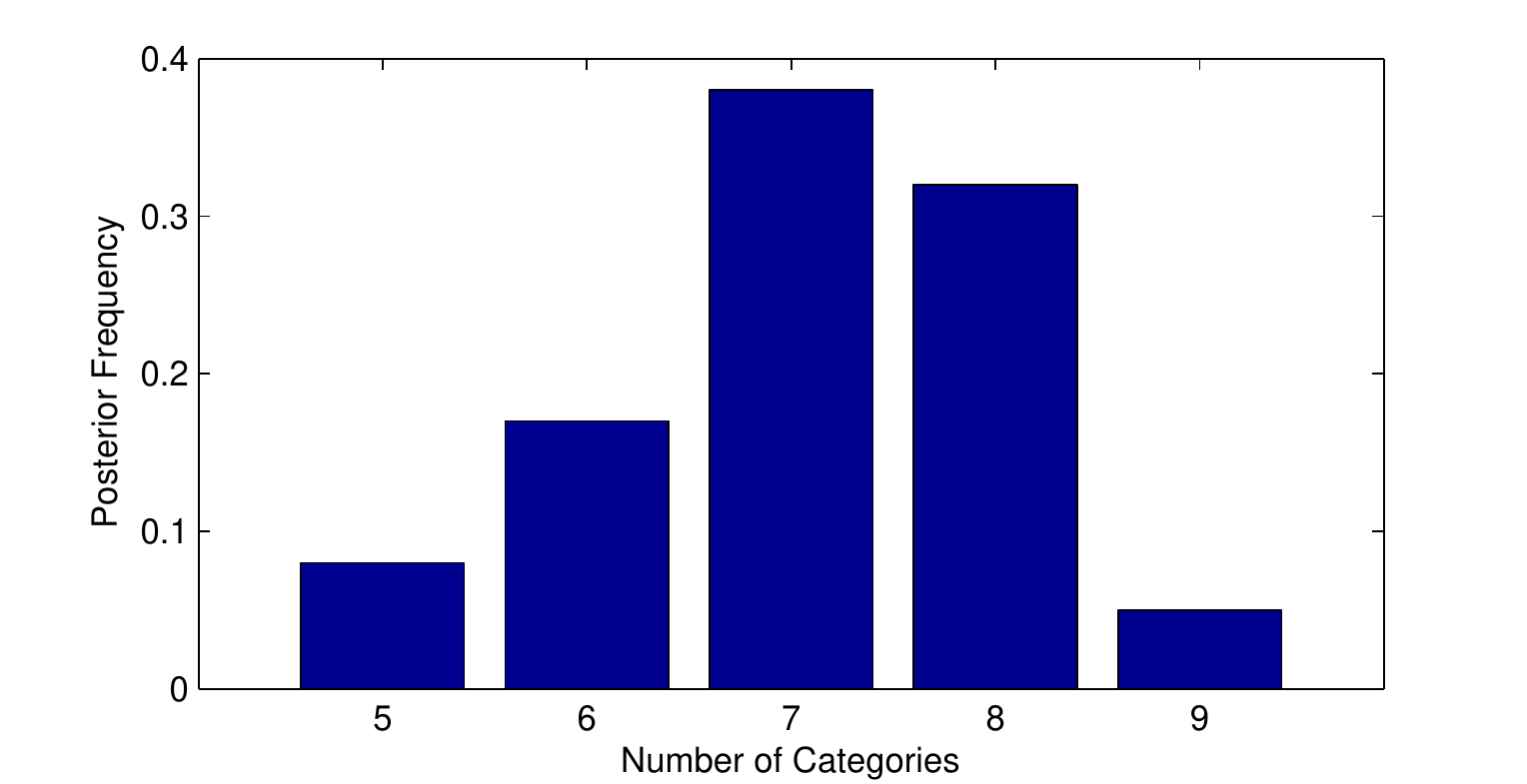}}}
\caption{Posterior frequencies of the inferred numbers of categories on TDT2 and ODP.}
\end{figure}

\subsection{Time efficiency}
The core sampling procedure of LBPL-NTM was implemented in C++, and all experiments were conducted in Matlab on a desktop with 3.60 GHz CPU. On the 4-way learning problems constructed from 20 Newsgroups, each round of our collapsed Gibbs sampling procedure takes about 0.9 second, which is a little slower than the speed of 0.7 second per sampling round of DLDA (implemented in C). We attribute this speed difference to the nonparametric nature of LBPL-NTM.

It is also observed empirically that both LBPL-NTM and DLDA run much faster than constraints based semi-supervised clustering methods. Besides, as mentioned above, SL and DPGM are quite inefficient for high dimensional text data, and PCA has to be used for them.

\vspace{0.05in}
\section{Conclusion and Future Work}
We proposed a nonparametric Bayesian method for learning beyond the predefined label space. Unlike existing methods which assume the number of unknown new categories in test data is known, our model can automatically infer this number via nonparametric Bayesian inference.
Empirical results show that: (a) compared with parametric approaches that use pre-specified true number of new categories, the proposed nonparametric approach yields comparable performance; and (b) when the exact number of new categories is unavailable, our approach has evident performance advantages.
Our model can be extended in several aspects, e.g., 1) adapt it to the online learning scenario with sequential Monte Carlo \cite{doucet2000sequential}; 2) explore multi-source text corpora with cross-domain learning \cite{zhuang2012mining,du2012multi}; and 3) leverage semantic representations such as attributes or class prototypes to bridge seen and unseen classes as in \cite{NIPS2018_7471}.

\smallskip
\section{Acknowledgments}
This work was supported by the National Natural Science Foundation of China (No. 61473273, 61602449, 61573335, 91546122, 61303059), Guangdong provincial science and technology plan projects (No. 2015B010109005), and the Science and Technology Funds of Guiyang (No. 201410012).

\smallskip

\bibliography{changyingdu}
\bibliographystyle{splncs03}
\end{document}